# A Formal Model of Dictionary Structure and Content

Nancy IDE
Dept. of Computer Science
Vassar College (USA)
ide@cs.vassar.edu

Adam KILGARRIFF
ITRI
University of Brighton (UK)
kilgarriff@itri.brighton.ac.uk

Laurent ROMARY
Equipe Langue et Dialogue
LORIA/CNRS (France)
romary@loria.fr

**Abstract.** We show that a general model of lexical information conforms to an abstract model that reflects the hierarchy of information found in a typical dictionary entry. We show that this model can be mapped into a well-formed XML document, and how the XSL transformation language can be used to implement a semantics defined over the abstract model to enable extraction and manipulation of the information in any format.

## 1 Introduction

The structure and content of dictionary entries and lexical information has been explored in considerable depth in the past (see, for instance [IDE ET AL. 1995]), primarily in order to determine a common model that can serve as a basis for encoding schemas and/or database formats. For the most part, descriptions of dictionary structure have been informed by the format of printed dictionaries, which varies considerably over dictionaries produced by different publishers and for different purposes, together with the requirements for instantiation in some encoding format (principally, SGML). However, the constraints imposed by these formats interfere with the development of a model that fully captures the underlying structure of lexical information. As a result, although schemas such as those provided in the *TEI Guidelines* exist, they do not provide a satisfactorily comprehensive and unique description of dictionary structure and content.

We believe that in order to develop a concrete and general model of dictionaries, it is essential to distinguish between the formal model itself and the encoding or database schema that may ultimately instantiate it. That is, it is necessary to consider, in the abstract, the form and content of lexical information independent of requirements and/or limitations imposed its ultimate representation as an encoded or printed object. This is especially important since these eventual representations will vary from one application to another; in particular, dictionaries may be encoded not only for the purposes of publishing in print or electronic form, but also for creating computational lexicons for use in natural language processing applications. It is therefore essential to develop a model that may be subsequently transformed into a variety of alternative formats.

In this paper, we outline a formal model for dictionaries that describes (a) the structure of dictionary information, (b) the information associated with this structure at various levels, and (c) a system of inheritance of information over this structure. We then show how the structure may be instantiated as a document encoded using the Extended Markup Language

(XML). Using the transformation language provided by the Extensible Style Language (XSL), we then demonstrate how the original XML instantiation may be transformed into other XML documents according to any desired configuration (including omission) of the elements in the original. Because of its generality, we believe our model may serve as a basis for representing, combining, and extracting information from not only dictionaries, but also terminology banks, computational lexicons, and, more generally, a wide variety of structured and semi-structured document types. An XML document type definition designed to be used with the formal framework described below is discussed in [ERJAVEC ET AL, 2000].

## 2 A formal representation of dictionaries

### 2.1 Basic notation

The underlying structure of a dictionary can be viewed as embedded partitions of a lexicon, in which no distinction is made among embedded levels (e.g., entry, homonym, sense, etc.).

A dictionary is thus a recursive structure comprised, at each level, of one or more nodes. This structure is most easily visualized as a tree, where each node may have zero or more children. That is, at any level $n$, a node is either a *leaf* (i.e., with no children) or can be decomposed as:

$$T=[T_1, T_2, ..., T_n]$$

where each $T_i$ is a node at level $n+1$.

Properties may be attached to any node in the dictionary structure with the *prop* predicate:

$$PROP(T,P)$$

indicates that the property P is attached to node T.

Properties are associated with nodes either by explicit assignment, or they may be *inherited* from the parent node. The object of our model is to identify the ways in which properties are propagated through levels of dictionary structure. For this purpose, we consider properties to be Feature-Value[1] pairs expressed as terms of the form FEAT(F,V), where F and V are tokens designating a *feature* (e.g., POS) and a *value*. In the simplest case, values are atomic (e.g., NOUN) but may also consist of sets of feature-value pairs. This representation is consistent with the base notation associated with feature structures [SHIEBER 1986], a common framework for representing linguistic information.

### 2.2 Propagating information across levels of dictionary structure

We define three types of features:

- *Cumulative features* that may take more than one value and may be thus inherited and combined along the dictionary structure. For example, for a cumulative feature DOMAIN, if the property FEAT(DOMAIN,NAVIGATION) is associated with a node at level $n$ and FEAT(DOMAIN,LAW) is associated with its child at level *n+1*, by inheritance the node at level *n+1* will be assigned the property FEAT(DOMAIN,NAVIGATION + LAW).

- *Overwriting features* that take only one value at a time. This implies that only one instance of an overwriting feature may appear at a given node and that the corresponding properties is propagated along the dictionary structure unless and until a new value is specified for that feature. In such a case, the new value "overwrites" the earlier one and is subsequently propagated to nodes in its subtrees.

- *Local features,* which apply only at the node with which they are associated; i.e., they are not propagated through the structure. Cross-references are an example of a local feature, since they apply only to the level of description with which they are directly associated.

Cumulative, overwriting, and local features are identified using predicates (CUM($F_C$), OVER($F_O$), and LOC($F_O$), respectively.

### 2.2.1 Basic notation

To model the behavior associated with cumulative and overwriting features, we consider two types of inferences:

1. Classical implication: $\varphi \rightarrow \psi$ which is monotonic

2. Defeasible implication: $\varphi > \psi$ where $\psi$ may be deduced from $\varphi$ unless there exists some information incompatible with $\psi$[2]

### 2.2.2 Propagating cumulative features

Since cumulative features may take several values at a time, their behavior can be described with a simple propagation rule:

$$PROP(T,FEAT(F_C,V)) \wedge CUM(F_C) \rightarrow \forall I \ PROP(T_I, FEAT(F_C,V)), T=[T_1, T_2, ..., T_N]$$

No specific constraints hold for cumulative features.

### 2.2.3 Propagating overwriting features

The propagation of an overwriting feature can be represented with a *defeasible* inference:

$$PROP(T,FEAT(F_O,V)) \wedge OVER(F_O) > \forall I \ PROP(T_I, FEAT(F_O,V)), T=[T_1, T_2, ..., T_N]$$

This inference is associated with a rule that expresses the incompatibility of two values of the overwriting feature at a given node:

$$PROP(T,FEAT(F_O,V)) \wedge OVER(F_O) \rightarrow \forall V' \neq V, \neg PROP(T,FEAT(F_O,V'))$$

### 2.2.4 Expressing dependencies among features

In certain cases, overwriting the value of a given feature has implications for the values of other features, and even for the existence of other features themselves. This typically occurs for *classes* of features; for example, for the class of features representing grammatical information, overwriting POS=N with POS=V implies that features such as GEN=F,

PLURALFORM=-S no longer apply, and that features such as PASTPART are now applicable. Here, we see that within the features supplying grammatical information, POS is a *privileged feature* because it dictates the features and values for other members of its class.

To deal with this situation it is necessary to introduce *dependency rules* that block the propagation of subordinate features when the privileged feature is overwritten. The following rule indicates that feature F is dependent on a given value V for feature G:

$$\text{PROP(T,FEAT(F,\_))} \rightarrow \text{PROP(T,FEAT(G,V)) dictionary}$$

For instance, to express the dependency between GEN and CAT=NOUN, the rule is

$$\text{PROP(T,FEAT(GEN,\_))} \rightarrow \text{PROP(T,FEAT(CAT,"N"))}$$

Rules of this kind apply only to overwriting features, since they block the activation of a defeasible rule.

## 3  Creating representations from dictionary entries

Lexical information in dictionary entries can be represented as a tree structure reflecting, in large part, the natural hierarchical organization of entries in printed dictionaries. This hierarchical organization (e.g., division into homographs, senses, sub-senses, etc.) enables information to be applied over all sub-levels in the hierarchy, thus eliminating the need to re-specify common information (for a detailed discussion, see [IDE, ET AL., 1995]; [IDE/VERONIS, 1995).

For example, consider the following definition from the *Collins English Dictionary (CED)*:

**EX. 1: disproof**

> **disproof** (dIs'pru:f)  *n.* **1.** facts that disprove something. **2.** the act of disproving.

The information in this entry can be represented in tree form as follows:

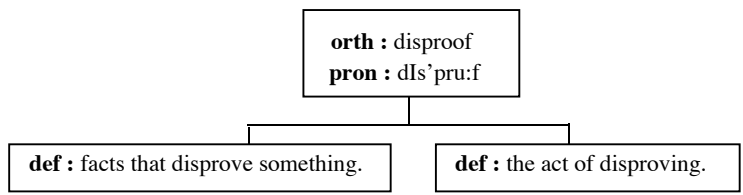

Each node in the tree represents a partition of the information in the entry, and information is inherited over sub-trees as defined in section 2.2. Thus in this example, orthographic form, pronunciation, and part-of-speech are inherited by both sub-nodes, each of which represents a partition of the information into two distinct (hence, sibling) senses. The following entry shows a different organization:

**EX. 2 : overdress** *(CED)*

> **overdress** *vb.* (***pron1***) 1. **To dress (oneself or another) too elaborately or finely. ~*n*.**
> (***pron2***) 2. **A dress that may be worn over a jumper, blouse, etc.**

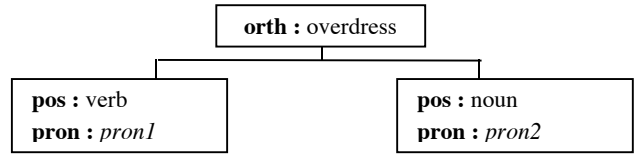

The orthographic form "overdress" appears at the top node and applies to the entire entry; the entry is then partitioned into two sub-trees, for verb and noun, each of which is associated with specific information about part of speech, pronunciation, and definition.[3]

**EX. 3 : gendarme** *(Le Petit Robert)*

> **gendarme** (**...**) *n.m.* (**XV°**; *gendarmes*; de *g e n s*, et *arme*)…II. (**1790**). *Mod.* **Militaire**
> **appartenant à un corps spécialement chargé de veiller au maintien de l'ordre et de la**
> **sureté de la publique…**(V. **Gendarmerie, Marechaussée**). *Brigade de gendarmes* (V.
> **Brigadier**). *Etre arrêté par les gendarmes.  Jouer au gendarme et au voleur.*  ◊ **Le**
> **gendarme : symbole de la force publique, de l'ordre.** *La peur du gendarme…*

The tree representing the information in this entry is:

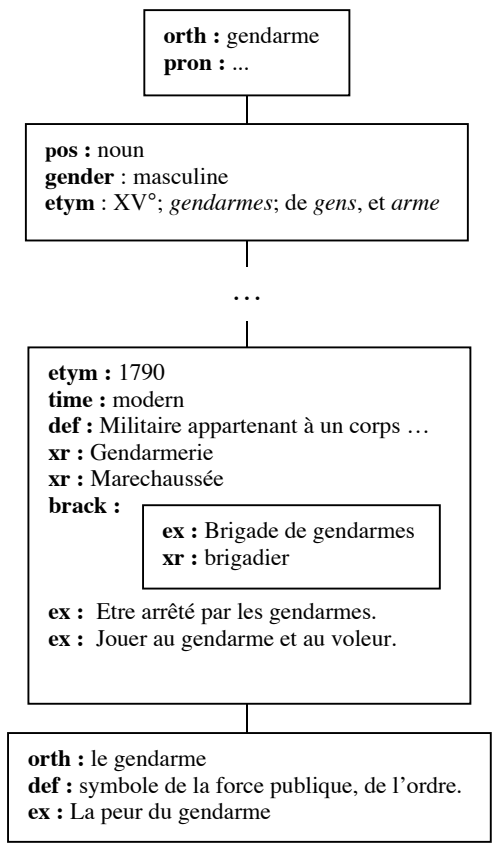

The value of the **brack** feature is itself a set of feature-value pairs (see section 2.1); the use of this generic feature provides a means to group **"Brigade de gendarmes"** and the cross-reference to "brigadier" together. Again, orthographic form and pronunciation appear at the highest level and apply to the entire entry, except at the lowest level, where the value of **orth** is overwritten for the subentry for "le gendarme".

The specification of alternatives, as for the two plural forms in Example 4, is common in dictionary entries when both alternatives apply to information lower in the hierarchy. However, this notation is effectively a shorthand indicating a new level of partitioning of the information, as shown in the corresponding tree structure:

**EX. 4 : pinna** *(CED)*

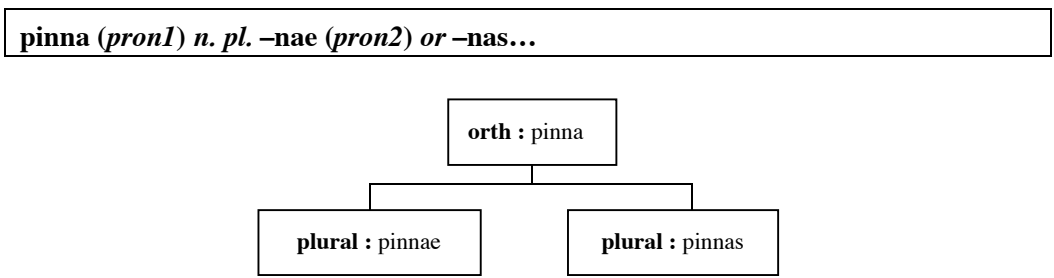

As it would in a dictionary entry, this results in the duplication of potentially large amounts of information. In section 5 we will show that the encoded version of the information avoids this duplication.

In general, the following rules apply to the creation of trees from dictionary entries:

- Headword information and other information appearing in the entry at the same level (typically, other form-related information such as pronunciation, hyphenation, etc.) is associated with the root.

- Sibling sub-trees for the current node (initially, the root) are introduced upon encountering the following:

- Partitions of the entry into different parts of speech;

- Partitions of the entry into senses at the same level (applied recursively to sub-senses);

- Partitions between the main entry and a related entry.

Although expressed informally here, a precise algorithm for tree creation can, in principle, be implemented, thereby enabling the automatic or semi-automatic creation of trees from dictionary entries in any of a variety of electronic forms (typesetter tapes, word processor output, etc.).

## 4    Extracting information from the tree

We define a *tree traversal* as any path starting at the root of the tree and following, at each node, a single child of that node. A *full traversal* is a path from the root to any leaf; a *partial traversal* extends from the root to any node in one of its subtrees.

As a tree created from a dictionary entry is traversed, each node is associated with a set of features including: (a) features associated with the node during tree creation, and (b) features determined by applying the rules for propagating overwriting, cumulative, and local features outlined in section 2.2. Thus, at any node, all applicable information is available for some unique partition of the lexical space. Nodes near the top of the tree represent very broad categories of partition; leaf nodes are associated with information for the most specific usage of the entry.

A traversal of the tree for Example 3 above *(gendarme)* proceeds as follows:

| | | |
|---|---|---|
| Root: | **orth :** | gendarme |
| | **pron :** | ... |
| Second: | **orth :** | gendarme |
| | **pron :** | ... |
| | **pos :** | noun |
| | **gen :** | masculine |
| | **etym :** | XV°; *gendarmes*; de *gens*, et *arme* |
| Third: | **orth :** | gendarme |
| | **pron :** | ... |
| | **pos :** | noun |
| | **gen :** | masculine |
| | **etym :** | 1790   // new value for etym overwrites value specified at level 2 |
| | **time :** | modern |
| | **def :** | Militaire appartenant à un corps … |
| | **xr :** | Gendarmerie   //xr and ex are local; no propagation |
| | **xr :** | Marechaussée |
| | **brack :** | [ **ex :**  Brigade de gendarmes |
| | | **xr :** brigadier] |
| Leaf: | **pron :** | ... |
| | **pos :** | noun |
| | **gen :** | masculine |
| | **etym :** | 1790 |
| | **time :** | modern |
| | **def :** | Militaire appartenant à un corps … |
| | **orth :** | le gendarme    // new value for orth overwrites value specified at root |
| | **def :** | symbole de la force publique, de l'ordre. |
| | **ex :** | La peur du gendarme |

The features *orth* (orthography) and *etym* (etymology) are overwriting features, and therefore values specified for these features at lower nodes of the tree replace those appearing above them. The feature *def* (definition) is cumulative, and so all definitions are retained during the root-to-leaf traversal. The features *ex* (example) and *xr* (cross-reference) are local and are therefore not propagated to lower nodes in the tree.

For "overdress", two traversals are possible:

(1) Root:          **orth :**  overdress
    Second level:  **orth :**  overdress
                   **pos :**   verb
                   **pron :**  *pron1*
                   **def:**    To dress (oneself or another) too elaborately or finely
(2) Root:          **orth :**  overdress
    Second level:  **orth :**  overdress
                   **pos :**   noun
                   **pron :**  *pron2*
                   **def :**   A dress that may be worn over a jumper, blouse, etc.

## 5  Encoding the information in XML

We define an XML encoding format for the structures described above:

**Elements**

- *<struc>* represents a node in the tree. <struc> elements may be recursively nested at any level to reflect the structure of the corresponding tree. <struc> is the *only* element in the encoding scheme that corresponds to the tree structure; all other elements provide information associated with a specific node (i.e., the node corresponding to the immediately enclosing <struc> element).[4]

- *<alt>* alternatives are bracketed in parallel <alt> elements, which may appear within any <struc>. The use of this element to encode alternatives corresponds to the shorthand often used in dictionary entries, where two equally applicable sets of information apply to the entire sub-tree.

- *<brack>* is a general-purpose bracketing element to group associated features.

- Base elements corresponding to features, including **orth, pron, hyph, syll, stress, pos, gen, case, number, gram, tns, mood, usg, time, register, geo, domain, style, def, eg, etym, xr, trans, and itype,** analogous to dictionary elements defined in the *TEI Guidelines*.

**Attributes**

Attributes are used to provide information specific to the element on which they appear and are not inherited in a tree traversal.

The following show the corresponding XML encoding for Examples 1, 2, and 3:

```
<struc>
    <orth>overdress</orth>
    <struc>
            <pos>verb</pos>
            <pron>...</pron>
            <def> To dress (oneself or another) too elaborately or finely </def>
    </struc>
     <struc>
            <pos>noun</pos>
            <pron>...</pron>
            <def> A dress that may be worn over a jumper, blouse, etc.</ def>
    </struc></struc>
```

**Figure 2. XML encoding for "overdress"**

```
<struc>
    <orth>gendarme</orth>
     < pron>...</pron>
    <struc>
            <pos>noun</pos>
            <gender>masc</gender>
            <etym> XV°; gendarmes; de gens, et arme</etym>
…
             <struc>
                    <etym>1790</etym>
                    <time>modern</time>
                    <def>Militaire appartenant à un corps …</def>
                    <xr type="see">Gendarmerie</xr>
                    <xr type="see">Marechaussée </xr>
                    <brack>
                            <ex>Brigade de gendarmes </ex>
                            <xr type="see">brigadier</xr></brack>
                    <ex>Etre arrêté par les gendarmes. </ex>
                    <ex>Jouer au gendarme et au voleur. </ex>
                        <struc>
                                <orth>le gendarme</orth>
                                <def>symbole de la force publique, de l'ordre.</def>
                                <ex>La peur du gendarme</ex>
                        </struc></struc></struc></struc>
```

**Figure 3. XML encoding for "gendarme"**

```
<struc>
    <orth>pinna</orth>
    < pron>pron1</pron>
    <pos>noun</pos>
    <struc>
            <alt>
                    <plural>pinnae</plural>
                    <pron>pron2</pron></alt>
            <alt>
                    < plural>pinnas </plural></struc>…
```

**Figure 4. XML encoding for first part of the entry "pinna"**

## 6    Transforming the XML document

The Extensible Style Language (XSL) is a part of the XML framework that enables transformation of XML documents into other XML documents. The best-known use of XSL is the formatting of documents for display on web browsers. However, XSL also provides a powerful transformation language that can be used to convert an XML document describing dictionary entries by selecting, rearranging, and adding information to it, using a powerful language enabling tree-traversal [CLARK, 1999]. Thus, a document encoded according to the specifications outlined in the previous section can be manipulated to serve any application that relies on part or all of its contents.

The XSL transformation language is relatively complex and will not be described in detail here. A short example can provide some idea of the possibilities. For example, the XSL script in Figure 5, applied to the XML-encoded entry for "overdress" given in Figure 2, produces an HTML document that will display the table in Figure 6 on a web browser:

```
<xsl:template match= "/">
  <html>
    <body>
     <table>
       <th>Word</th>
       <th>PoS</th>
       <th>Meaning</th>
       <xsl:foreach select="struc[pos]"/>
          <tr>
            <td><xsl:value-of select="ancestor::struc[orth]"/></td>
            <td><xsl:value-of select="//pos"/></td>
            <xsl:foreach select="//def"/>
               <td><xsl:value-of select="."/></td>
            </xsl:foreach>
          </tr>
       </xsl:foreach>
     </table>
    </body>
  </html>
</xsl:template>
```

**Figure 5. XSL code to produce an HTML document**

| Word | PoS | Meaning |
|------|-----|---------|
| overdress | verb | To dress (oneself or another) too elaborately or finely |
| overdress | noun | A dress that may be worn over a jumper, blouse, etc. |

**Figure 6. Display of HTML output for XSL code in Figure 5**

XSL's transformation language can also be used to implement the semantics of inheritance and overwriting outlined in section 2. The XSL code in Figure 7 shows the implementation for overwriting and cumulative features. The result is another XML document, but with the inherited information explicitly represented, as shown in Figure 8.

```
<xsl:stylesheet version="1.0" xmlns:xsl="http://www.w3.org/1999/XSL/Transform">

<xsl:template match="/">
<xsl:apply-templates/>
</xsl:template>

<!-- the recursive copy is initiated at the dict level -->
<xsl:template match="dict">
  <dict>
        <xsl:apply-templates/>
  </dict>
</xsl:template>

<!-- each struc is copied with supplementary features depending on 'over' or 'cum'
rules -->
<xsl:template match="struc">
  <struc>
        <!-- rule for an overwriting feature (here, "orth") -->
        <!-- if there is not already an 'orth'  take the first one you encounter -->
        <xsl:if test="not(orth)">
              <xsl:copy-of select="ancestor::node()[orth][1]/orth"/>
        </xsl:if>

        <!-- rule for a cumulative feature (here, "ex") -->
        <!-- copy all the 'ex' you find above in the tree -->
        <xsl:copy-of select="ancestor::node()/ex"/>

        <xsl:apply-templates/>
        </struc>
</xsl:template>

<!-- any other node is just copied -->
<xsl:template match="*">
<xsl:copy-of select="."/>
</xsl:template>

</xsl:stylesheet>
```

**Figure 7. XSLT code implementing overwriting and cumulative inheritance**

```
<?xml version="1.0" encoding="utf-8"?>
<dict>
<struc>
    <orth>overdress</orth>
    <struc><orth>overdress</orth>
        <pos>verb</pos>
        <pron>zzzz</pron>
        <def> To dress (oneself or another) too elaborately or finely </def>
    </struc>
    <struc><orth>overdress</orth>
        <pos>noun</pos>
        <pron>yyyy</pron>
        <def> A dress that may be worn over a jumper, blouse, etc.</def>
    </struc>
</struc>
</dict>
```

**Figure 8. Results of XSLT code in Figure 7 for "overdress"**

# 7   Conclusion

This brief overview is intended to show that a general model of lexical information conforms to an abstract model that reflects the hierarchy of information found in a typical dictionary entry, and that this model can be mapped into a well-formed XML document.

The XSL transformation language can be used to implement a semantics defined over the abstract model, thus enabling sophisticated query and manipulation of the data to serve a variety of ends.

## 8    Acknowledgements

The work reported in this paper was supported in part by National Science Foundation Grant INT-9815830, the Centre National de la Recherche Scientifique (France), and the European Commission INCO-COPERNICUS project CONCEDE (no. PL96-1142).

## 9    Notes

[1]  We deliberately avoid the term "attribute" to avoid confusion with the SGML/XML vocabulary.

[2] At this stage, our notation is independent of any specific theory (default logic, autoepistemic logic, circumscription, conditional logics etc.).

[3] Note that although the presentation of the verb and noun senses of the headword is effectively linear in the printed form of entry, the reader understands the implied parallelism apparent in the derived structure.

[4] XML documents are also described as trees, where the "parent" of a given element is the element in which it is immediately enclosed. To avoid confusion, we use the term "tree" and the associated terminology to refer only to the structures outlined in section 4.

## 10   References

Clark, James (ed.) (1999). XSL Transformations (XSLT). Version 1.0. W3C Recommendation.http://www.w3.org/TR/xslt.

Erjavec, Tomaz, Evans, Roger, Ide, Nancy, Kilgarriff, Adam (2000). "The CONCEDE model for Lexical Databases", in *Proceedings of the Second International Language Resources and Evaluation Conference*, to appear.

Ide, Nancy, Le Maitre, Jacques, Véronis, Jean (1995). "Outline of a Model for Lexical Databases", in *Current Issues in Computational Linguistics: In Honour of Don Walker*. *Linguistica Computazionale* IX, X (Pisa), pp. 283-320. [reprinted from *Information Processing and Management,* 29, 2, pp. 159-186].

Ide, Nancy, Véronis, Jean (1995). "Encoding dictionaries", in Ide, N., Veronis, J. (Eds.) *The Text Encoding Initiative: Background and Context*. Kluwer Academic Publishers, Dordrecht, pp. 167-80.

Shieber, Stephen (1986). *An Introduction to Unification-based Approaches to Grammar*. CSLI Lecture Notes Series, Chicago: University of Chicago Press.